\pdfoutput=1

\documentclass[11pt]{article}

\usepackage[]{acl}

\usepackage{times}
\usepackage{latexsym}

\usepackage[T1]{fontenc}

\usepackage[utf8]{inputenc}

\usepackage{microtype}

\usepackage{graphicx}

\title{Data-Efficiency with a Single GPU: An Exploration of Transfer Methods for Small Language Models} 

\author{
    Alon Albalak\textsuperscript{1},
    Akshat Shrivastava\textsuperscript{2},
    Chinnadhurai Sankar\textsuperscript{2},
    Adithya Sagar\textsuperscript{2},
    Mike Ross\textsuperscript{2} \\
    \textsuperscript{1}University of California, Santa Barbara\qquad \textsuperscript{2}Meta AI \\
    \texttt{alon\_albalak@ucsb.edu}
    }

\begin{document}
\maketitle
\begin{abstract}
Multi-task learning (MTL), instruction tuning, and prompting have recently been shown to improve the generalizability of large language models to new tasks. However, the benefits of such methods are less well-documented in smaller language models, with some studies finding contradictory results. In this work, we explore and isolate the effects of (i) model size, (ii) general purpose MTL, (iii) in-domain MTL, (iv) instruction tuning, and (v) few-shot fine-tuning for models with fewer than 500 million parameters. Our experiments in the zero-shot setting demonstrate that models gain 31\% relative improvement, on average, from general purpose MTL, with an additional 37.6\% relative gain from in-domain MTL. Contradictory to prior works on large models, we find that instruction tuning provides a modest 2\% performance improvement for small models.
\end{abstract}

\section{Introduction}
Many recent works have demonstrated the benefits of prompting for large language models (see \citet{prompt_survey} for an extensive survey). While prompts started as simple task identifiers (eg. "topic" for topic classification) they have expanded to include answer templates, examples, and instructions \cite{2020t5, FETA,FLAN,Mishra2021NaturalIB,Ouyang2022TrainingLM}. Studies on utilizing prompts have shown that as model sizes scale up, the generalization abilities of a model increase \citep{gpt3, lester-etal-2021-power,demonstrations_icl}. However, utilizing models on the hundred-billion parameter scale is not accessible for most researchers and practitioners. Furthermore, \citet{wei2022emergent} show that trends for large language models do not hold for smaller language models. 
For this reason, \textit{it is crucial} that we must empirically find the trends that occur in smaller models and cannot rely on studies of larger models.

Interestingly, some findings on instruction tuning have been contradictory across studies. For example, \citet{FLAN} find that models with fewer than 8B parameters see decreases in generalization when utilizing instructions, whereas \citet{instructdial} find consistent gains in models with 3B and fewer parameters. To conflate these results further though, \citet{instructdial} only consider 2 situations: when inputs include instructions and answer templates, or neither.

Simultaneously with the emergence of prompting, the explicit multi-task learning (MTL) paradigm emerged, with works such as Muppet \citep{aghajanyan-etal-2021-muppet} or T0 \citep{sanh2022multitask} and their variants. Explicit MTL has been demonstrated as a means of improving the downstream performance of pre-trained language models in data-constrained settings. In this work we consider 2 types of MTL: \textbf{general purpose} and \textbf{in-domain}. Specifically, general purpose MTL consists of training across a wide variety of tasks and domains, whereas in-domain MTL consists of training across a variety of tasks that all occur within a domain.

One limitation of many previous works on prompting and multi-task learning is that they focus on language models in the billion-parameter scale. For situations with latency and memory limitations, small models may be the only option. In this work, we study an example of such a domain; dialogue.

In this work we bridge the gap between previous studies by exploring the effects of a variety of factors on the zero- and few-shot generalizability on modestly sized language models (<500M parameters). Specifically, we run experiments to find the effects of: (i) model size, (ii) general purpose MTL, (iii) in-domain MTL, (iv) instruction tuning, and (v) fine-tuning with and without instructions. Additionally, 
we perform a linguistic analysis on instruction wording and
analyze variations in performance across task instructions.


In this study, we find that
\textbf{(1)} In-domain multi-task learning (MTL) gives the largest improvements to generalizability, up to 80\% increased relative performance, and 37.6\% on average across all models
\textbf{(2)} Increasing model size alone has little effect on generalization, but when combined with in-domain MTL leads to double the (already strong) performance improvement of in-domain MTL
\textbf{(3)} General purpose MTL can provide large gains (57\% improvement) for downstream tasks which closely resemble the MTL tasks, but still provides modest gains (5\%) even for tasks which are more dissimilar
\textbf{(4)} Instruction tuning during in-domain MTL provides modest gains of just over 2\% performance, regardless of model size.
\section{Experiments}

\paragraph{Data}
For this study we utilize 46 tasks from the Instructdial dataset \citep{instructdial}. Each task is converted into a sequence-to-sequence format with an answer template, allowing a single generative model to perform all tasks. Each task contains 3 to 10 instructions, with 4.4 instructions on average.
For our zero-shot experiments, we use 3 splits of train/test tasks, where each split contains 40 training tasks and 6 test tasks.
For our few-shot experiments, we use the first data split only.
Tasks are divided into classification and generation, where classification tasks are evaluated on accuracy and generation tasks by Rouge-L scores. The full list of tasks and information on train/test splits can be found in the section \ref{sec:data} of the Appendix.

\paragraph{Models}
In our experiments, we utilize 3 variants of BART model \citep{Lewis2020BARTDS}: BART-Base, BART-Large and BART0++ \citep{Lin2022UnsupervisedCG}. BART0++ is a BART-Large that has been multi-task trained on PromptSource \citep{bach2022promptsource} in the same fashion as T0++ \citep{sanh2022multitask}.\footnote{All pre-trained models were downloaded from the HuggingFace Transformers library.}

\begin{figure}[t!]
    \centering
    
    \includegraphics[width=\columnwidth]{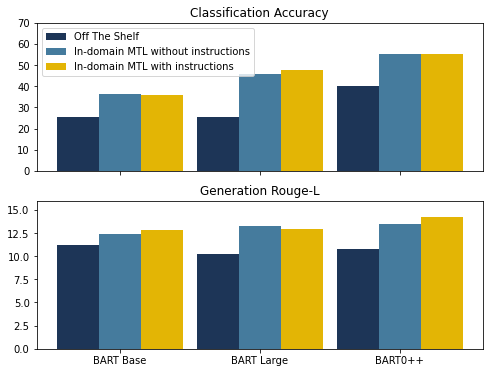}
    
    \caption{Average performance on 10 zero-shot classification tasks (top) and 8 zero-shot generation tasks (bottom) comparing pre-trained models (Off the shelf) with models explicitly multi-task trained on in-domain data with and without instructions.
    }
    \label{fig:aggregate_scores}
\end{figure}

\begin{figure*}[t]
    \centering
    \includegraphics[width=0.94\linewidth]{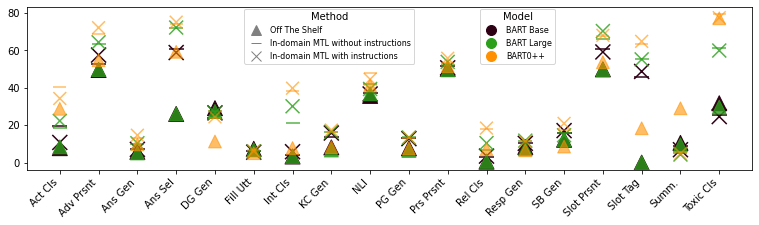}
    \caption{Absolute scores on 18 zero-shot tasks. Full task names are found in section \ref{sec:abbreviations} of the Appendix.}
    \label{fig:zeroshot_absolute_scores}
\end{figure*}

\paragraph{Experimental Setup}
To study the effects of (i) model size, (ii) general purpose MTL, (iii) in-domain MTL, and (iv) instruction tuning, we run a series of zero-shot experiments.
In order to measure the effect of (i) model size, we compare performance between BART-Base (139M parameters) and BART-Large (406M parameters).
To measure the effect of (ii) general purpose MTL, we compare performance between BART-Large and BART0++.
To study the effect of (iii) in-domain MTL, we compare each model trained with in-domain MTL against an off-the-shelf version that is directly tested on each split.
To measure the effect of (iv) in-domain MTL with instructions, we include instructions in addition to the answer template in the input sequences.
All experiments were repeated with 3 random seeds and reported scores are means.\footnote{Further training details on hyperparameters are in section \ref{sec:training_dets} of the Appendix}

In addition to zero-shot experiments, we also consider the situation where we have small quantities of data for fine-tuning. We run experiments with 10/10, 50/50, and 100/100 training/validation samples. For the few-shot experiments, we study the effects of (i) model size, (ii) general-purpose MTL, (iv) instruction tuning, and (v) fine-tuning with or without instructions.
\section{Findings}

Figure \ref{fig:aggregate_scores} shows the average zero-shot performance divided into classification and generation tasks. Figure \ref{fig:zeroshot_absolute_scores} shows the absolute scores for all models and methods on each of the 18 test tasks.

\paragraph{Effects of Model Size}
When comparing off-the-shelf versions of BART-Base and BART-Large, we find nearly identical performance across classification tasks, and slightly better performance for BART-Base (11.2 vs. 10.2 Rouge-L) on generation tasks. However, the benefits of model size are demonstrated once the models have been further trained using in-domain MTL (Figure \ref{fig:aggregate_scores}). We find that with in-domain MTL the base model improves its average score by 6.5. but the large model doubles that improvement, increasing it's score by 13.3 points averaged across all tasks.

\begin{figure*}[t]
    \centering
    \includegraphics[width=0.95\linewidth]{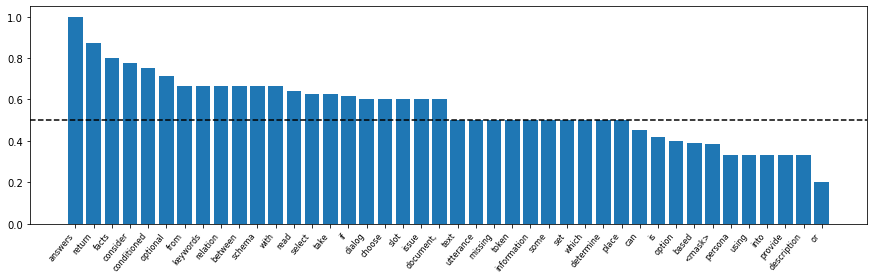}
    \caption{Percentage of occurrences of a word that lead to better than average performance for an instruction. Results calculated from BART-Base model and only includes words that occur in more than 5 instructions.}
    \label{fig:word_analysis_base}
\end{figure*}

\paragraph{Effects of General Purpose Multi-task Learning}
When comparing BART-Large and BART0++ we see improvements from general MTL on 14/18 tasks, and an average absolute improvement of 14.5 accuracy (57.1\% relative) on classification tasks, and more modest improvement of 0.6 Rouge-L (5\%) on generation tasks (Figure \ref{fig:aggregate_scores}). This large discrepancy is likely due to the distribution of tasks in the P3 dataset \cite{bach2022promptsource} used to train BART0++, which consists almost entirely of classification tasks with only summarization as a generation task. Figure \ref{fig:zeroshot_absolute_scores} shows that, indeed, an off-the-shelf BART0++ outperforms all other methods on summarization, including in-domain MTL.

\paragraph{Effects of In-Domain Multi-Task Learning}
We find that in-domain MTL (without instructions) contributes the largest portion to the final generalization ability of each model. As shown in Figure \ref{fig:aggregate_scores}, BART-Large gets the most improvement, with gains of 20.4 points in accuracy (80\% relative improvement) on classification tasks and 3 Rouge-L (29.3\%) for generation tasks. Bart-Base gets 41.8\% and 11.5\% relative improvements, and BART0++ gets 37.7\% and 25.3\% relative improvements. Collectively, this experiment and the previous experiments on general purpose MTL demonstrate the importance of matching both the domain and the task distribution during MTL to the downstream tasks and domain of interest. Additionally, as previously mentioned, in-domain MTL combined with the increased capacity of a larger model shows even greater improvements.

\begin{figure*}[t]
    \centering
    \includegraphics[width=0.95\linewidth]{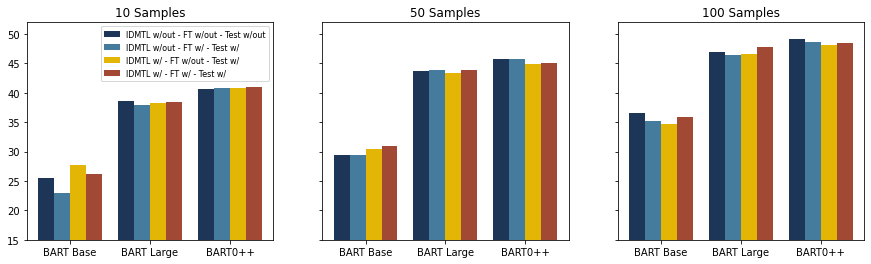}
    \caption{Average performance across 6 tasks when fine-tuning with 10, 50, or 100 samples. IDMTL refers to in-domain MTL and FT is fine-tuning. "w/" and "w/out" refer to with and without instructions.}
    \label{fig:fewshot_aggregate_scores}
\end{figure*}

\paragraph{Effects of Instruction Tuning on In-Domain Multi-Task Learning}
Next, we compare the performance of in-domain MTL with and without instructions. The benefits of instruction tuning on small models is less prominent than the three previous variables, but is still beneficial overall. Figure \ref{fig:aggregate_scores} shows that BART-Base improves by 3\% on generation tasks, but loses 1\% accuracy on classification. To the contrary, BART-Large improves by 4\% on classification tasks, and loses 2\% Rouge-L on generation tasks. Interestingly, BART0++ sees no difference in performance on classification tasks and improves by 5\% on generation tasks. These results run counter to those of \citet{FLAN}, which found that instruction tuning can degrade performance of models with fewer than 8B parameters by about 10\%. This may be partly due to the in-domain nature of the instructions utilized in our experiments (all instructions are related to dialogue), suggesting that future works on instruction tuning for small models should focus on (1) domain-specific wording used in instructions, and (2) expanding the number of domains included in instruction sets to see more general benefits.

\paragraph{Findings on Sensitivity to Instructions}
To better understand the importance of wording and draw insights, we take a closer look at the tasks which had highest variance across instructions. First, we find that Answer Selection is the task with highest variance (BART-Base scores range from 27.3 to 63) and find that the three worst performing instructions include variations of "select an option that can substitute <MASK>". The three instructions including this phrase average an accuracy of 39, while the remaining seven average an accuracy of 60.1. This large discrepancy is likely connected to the unnaturalness of the <mask> token, and that it is unlikely to have appeared in a similar context in the pre-training corpus, and only appears in 2/46 tasks in our in-domain dataset. The other task which utilizes the <mask> token is the "Fill-in the Missing Utterance" task, which also achieves very poor performance across all models and methods (with and without instructions). This is a strong reminder that to create generalizability in language models, it is crucial to match the downstream task to the pre-training data.

Next, we analyze individual instruction words which most frequently give better than mean performance (Figure \ref{fig:word_analysis_base}). Interestingly, we find that "return" (as used in "return a response to the conversation") almost always leads to better than average performance (7/8 occurances for BART-Base and Large, and 8/8 for BART0++), although it only occurs in 3 tasks, and 8 instructions.

Finally, we look at the standard deviation between instructions, averaged across all tasks and find very little difference between models, with slightly increasing variation as models get larger, and are pretrained (BART-Base: 0.848, BART-Large: 0.867, BART0++: 0.882). At first glance, this seems to suggest that BART-Base is most robust to wording in instructions, but this is more likely due to the smaller number of tasks which BART-Base can meaningfully perform, as seen in Figure \ref{fig:zeroshot_absolute_scores}.

\paragraph{Few-shot Experiments}
For our few-shot experiments we find that, similar to the zero-shot setting, the combination of increasing model size and additional training data is a large contributor to improved performance. On average, the relative improvement from BART-Base to Large is 50\%, 45.4\%, and 31.7\% for 10, 50, and 100 samples, respectively. On the other hand, the improvement due to general purpose MTL is only 6.6\%, 3.8\%, and 3.5\%. When considering our 3 methods of including instructions, we find that always including instructions is generally the best, outperforming the other 2 methods in 6/9 settings. We expected that as additional fine-tuning samples are added, the benefits of instruction-tuning will diminish due to the model learning an implicit mapping of the task. Indeed, we find that for BART-Base, training with instructions is beneficial with 50 or fewer fine-tuning samples, above which point training without any instructions gives best performance. Interestingly, BART0++ gains no performance from using instructions above 10 fine-tuning samples.
\section{Limitations}
While this study does it best to separately study each variable that contributes to performance, there are other variables that may also have significant impacts on generalization. For example, the BART-Base and Large models are both trained on the same dataset, and while this is crucial to determine the effect of model size isolated from other factors, it means that we have only run experiments with 1 pre-training corpus. Ideally, to account for the effect of pre-training corpus, we would train each of the BART models from scratch with a different pre-training corpus, but that may be quite costly.

Additionally, this study only considers a single model architecture: encoder-decoder. The same experiments can be run with a decoder-only model and may find different results. Running these experiments with a pre-trained decoder-only model would allow us to study both a different architecture and pre-training corpus, but would also confound the two variables.

In this work, we did not consider the situation where the in-domain dataset is included during general purpose MTL, which may have provided the best performance, based on previous works studying transfer learning \cite{FETA}.

Finally, while we focus on small models, because there is no BART model smaller than base, we were only able to go down to 139 million parameters. However, since the conclusion of our study, the OPT model has been released with many small sizes that would make for an even more in-depth study.

\bibliography{references}
\bibliographystyle{acl_natbib}

\appendix
\section{Training/Testing Data}
\label{sec:data}

The full list of tasks is:

Act Classification, Act Generation, Advice Generation, Advice Present, Answer Generation, Answer Selection, Begins-with Controlled Generation, Belief State Generation, Count Response Words, Database-based Generation, Deal Present, Dialfact Classification, Document Grounded Generation, Edit Generation, Emotion Generation, Emotion Tagging, Ends-with Controlled Generation, Evaluation-Binary, Evaluation-Ranking, Fill-in the Missing Utterance, Find the Incoherent Utterance, Graph-based Generation, Intent Classification, Intent Present, Keyword Controlled Generation, Knowledge Grounded Generation, Natural Language Inference, Non-Toxic Feedback Generation, Persona Grounded Generation, Persuasion Generation, Persuasion Present, Persuasion Strategy, Question Generation, Recovery Generation, Relation Classification, Relation Present, Response Generation with n Words, Response Generation, Schema-based Generation, Slot Present, Slot Tagging, Slot-Value Generation, Summarization, Target Controlled Generation, Toxic Response Classification.

The first data split uses the following 6 tasks for testing, and the remainder for training:
\begin{itemize}
    \item Act Classification
    \item Document Grounded Generation
    \item Intent Classification
    \item Keyword Controlled Generation
    \item Relation Classification
    \item Slot Present
\end{itemize}

The second data split uses the following 6 tasks for testing, and the remainder for training:
\begin{itemize}
    \item Answer Selection
    \item Natural Language Inference
    \item Persuasion Present
    \item Schema-based Generation
    \item Slot Tagging
    \item Summarization
\end{itemize}

The third data split uses the following 6 tasks for testing, and the remainder for training:
\begin{itemize}
    \item Advice Present
    \item Answer Generation
    \item Fill-in the Missing Utterance
    \item Persona Grounded Generation
    \item Response Generation
    \item Toxic Response Classification
\end{itemize}

\section{Task Abbreviations}
\label{sec:abbreviations}
\begin{itemize}
    \item Act Cls - Act Classification
    \item Adv Prsnt - Advice Present
    \item Ans Gen - Answer Generation
    \item Ans Sel - Answer Selection
    \item DG Gen - Document Grounded Generation
    \item Fill Utt - Fill-in the Missing Utterance
    \item Int Cls - Intent Classification
    \item KC Gen - Keyword Controlled Generation
    \item NLI - Natural Language Inference
    \item PG Gen - Persona Grounded Generation
    \item Prs Prsnt - Persuasion Present
    \item Rel Cls - Relation Classification
    \item Resp Gen - Response Generation
    \item SB Gen - Schema-based Generation
    \item Slot Prsnt - Slot Present
    \item Slot Tag - Slot Tagging
    \item Summ. - Summarization
    \item Toxic Cls - Toxic Response Classification
\end{itemize}

\section{Training Details}
\label{sec:training_dets}
We train all models for a maximum of 3 epochs, and utilize validation based early stopping. To determine the learning rate, we trained each model on a single seed and validate the best learning rate in \{1e-5, 5e-5, 1e-4\}, then train for 2 additional seeds using the best learning rate. We found for all models that 5e-5 was the best learning rate. For all experiments we use the AdamW optimizer.

\section{Additional Figures}



\begin{figure*}[t]
    \centering
    \includegraphics[width=\linewidth]{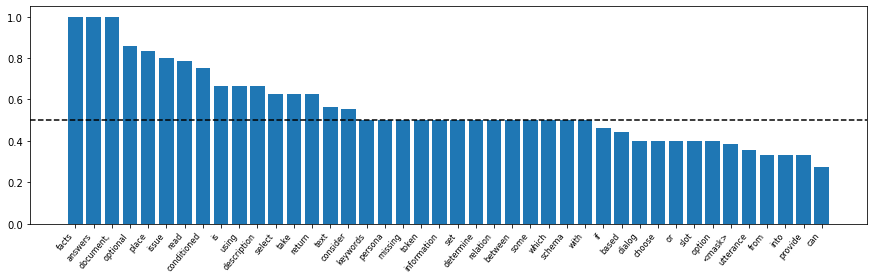}
    \caption{Percentage of occurrences of a word that lead to better than average performance for an instruction. Results calculated from BART-Large model and only includes words that occur is more than 5 instructions.}
    \label{fig:word_analysis_large}
\end{figure*}

\begin{figure*}[t]
    \centering
    \includegraphics[width=\linewidth]{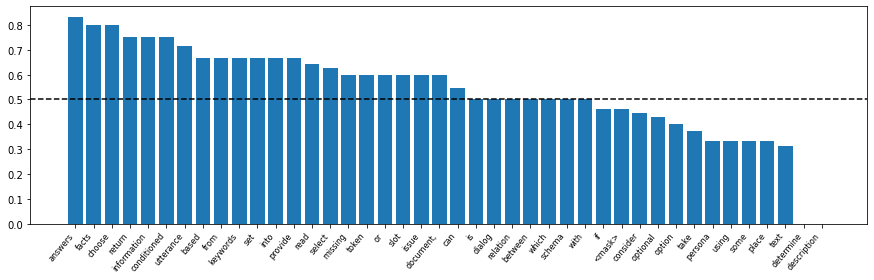}
    \caption{Percentage of occurrences of a word that lead to better than average performance for an instruction. Results calculated from BART0++ model and only includes words that occur is more than 5 instructions.}
    \label{fig:word_analysis_b0}
\end{figure*}

\begin{figure*}[t]
    \centering
    
    \includegraphics[width=\linewidth]{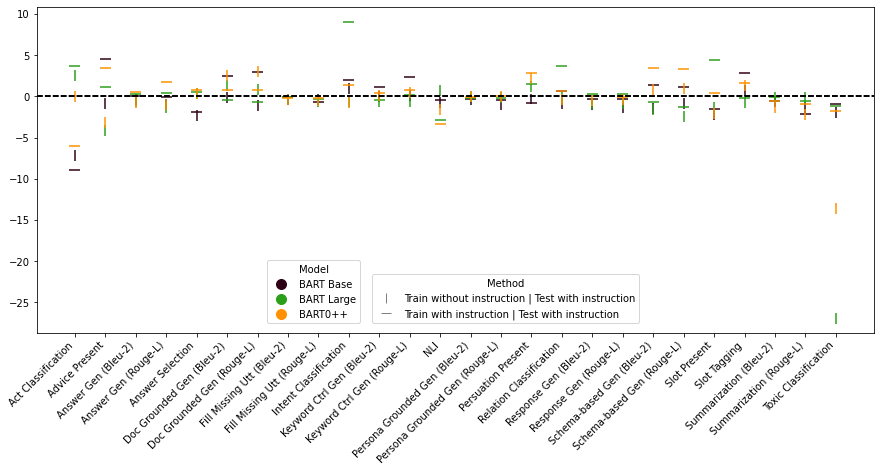}
    
    \caption{Performance comparison against a baseline model that is trained without instructions. Values shown are the absolute score difference between the designated method and the baseline model.}
    \label{fig:zeroshot}
\end{figure*}


\begin{figure*}[t]
    \centering
    
    \includegraphics[width=\linewidth]{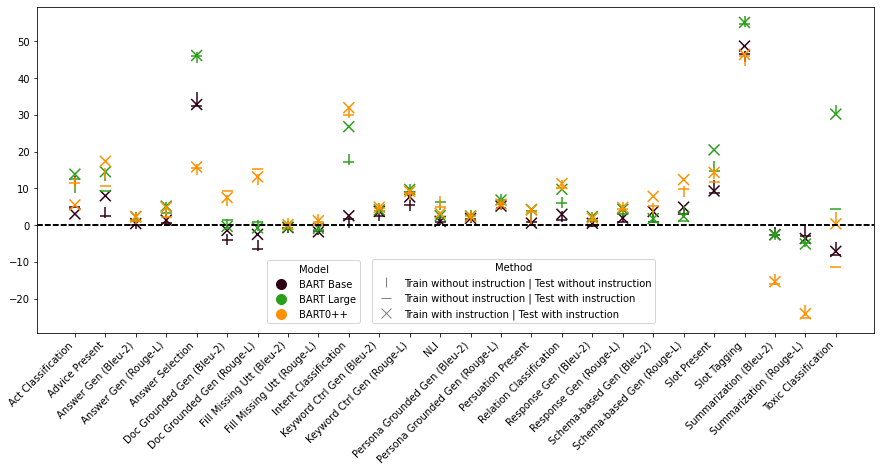}
    
    \caption{Performance comparison against an off-the-shelf baseline model. Values shown are the absolute score difference between the designated method and the baseline model.}
    \label{fig:zeroshot_pretrained_baseline}
\end{figure*}


\end{document}